\documentclass[letterpaper, 10 pt, conference]{ieeeconf}  

\IEEEoverridecommandlockouts                              

\usepackage{graphics} 
\usepackage{epsfig} 
\usepackage{amsmath} 
\usepackage{hyperref}
\usepackage{xcolor}
\usepackage{tikz}
\usetikzlibrary{calc}
\usetikzlibrary{arrows}
\usepackage{pgfplots}
\usepackage{siunitx}

\title{\LARGE \bf
DA-LMR: A Robust Lane Marking Representation for Data Association
}

\author{Miguel Ángel Muñoz-Bañón$^{1}$,  Jan-Hendrik Pauls$^{2}$,  Haohao Hu$^{2}$ and Christoph Stiller$^{2}$
\thanks{This work has been supported by the regional Valencian Community Goverment and the European Regional Development Fund (ERDF) through the grants ACIF/2019/088 and BEFPI/2021/069.}
\thanks{$^{1}$Author is with the Group of Automation, Robotics and Computer Vision (AUROVA), University of Alicante, San Vicente del Raspeig S/N, Alicante, Spain.
        {\tt\small miguelangel.munoz@ua.es}}%
\thanks{$^{2}$Authors are with Institute of Measurement and Control Systems, Karlsruhe Institute of Technology, Karlsruhe, Germany.}%
}

\begin{document}

\maketitle
\thispagestyle{empty}
\pagestyle{empty}

\begin{abstract}
While complete localization approaches are widely studied in the literature, their data association and data representation subprocesses usually go unnoticed. However, both are a key part of the final pose estimation.

In this work, we present DA-LMR (Delta-Angle Lane Marking Representation), a robust data representation in the context of localization approaches. We propose a representation of lane markings that encodes how a curve changes in each point and includes this information in an additional dimension, thus providing a more detailed geometric structure description of the data. We also propose DC-SAC (Distance-Compatible Sample Consensus), a data association method. This is a heuristic version of RANSAC that dramatically reduces the hypothesis space by distance compatibility restrictions.

We compare the presented methods with some state-of-the-art data representation and data association approaches in different noisy scenarios. The DA-LMR and DC-SAC produce the most promising combination among those compared, reaching 98.1\% in precision and 99.7\% in recall for noisy data with 0.5 m of standard deviation.

\end{abstract}

\section{INTRODUCTION}
Localization is an essential part of autonomous driving. For past years, there has been a wide variety of works in different approaches like \textit{Simultaneous Localization and Mapping} (SLAM) \cite{cadena2016past}, localization in previously built HD maps \cite{pauls2020monocular}, and geo-referencing using aerial imagery \cite{hu2019accurate}. A data association process becomes necessary for all these approaches to find correspondences between landmarks in maps and detections from the onboard sensors. The typical probabilistic models used in localization, such as factor-graphs or Bayesian filters, need these correspondences for inference. Hence, these associations are crucial for pose estimation.

\begin{figure}[t]
\centering
\includegraphics[width=220pt]{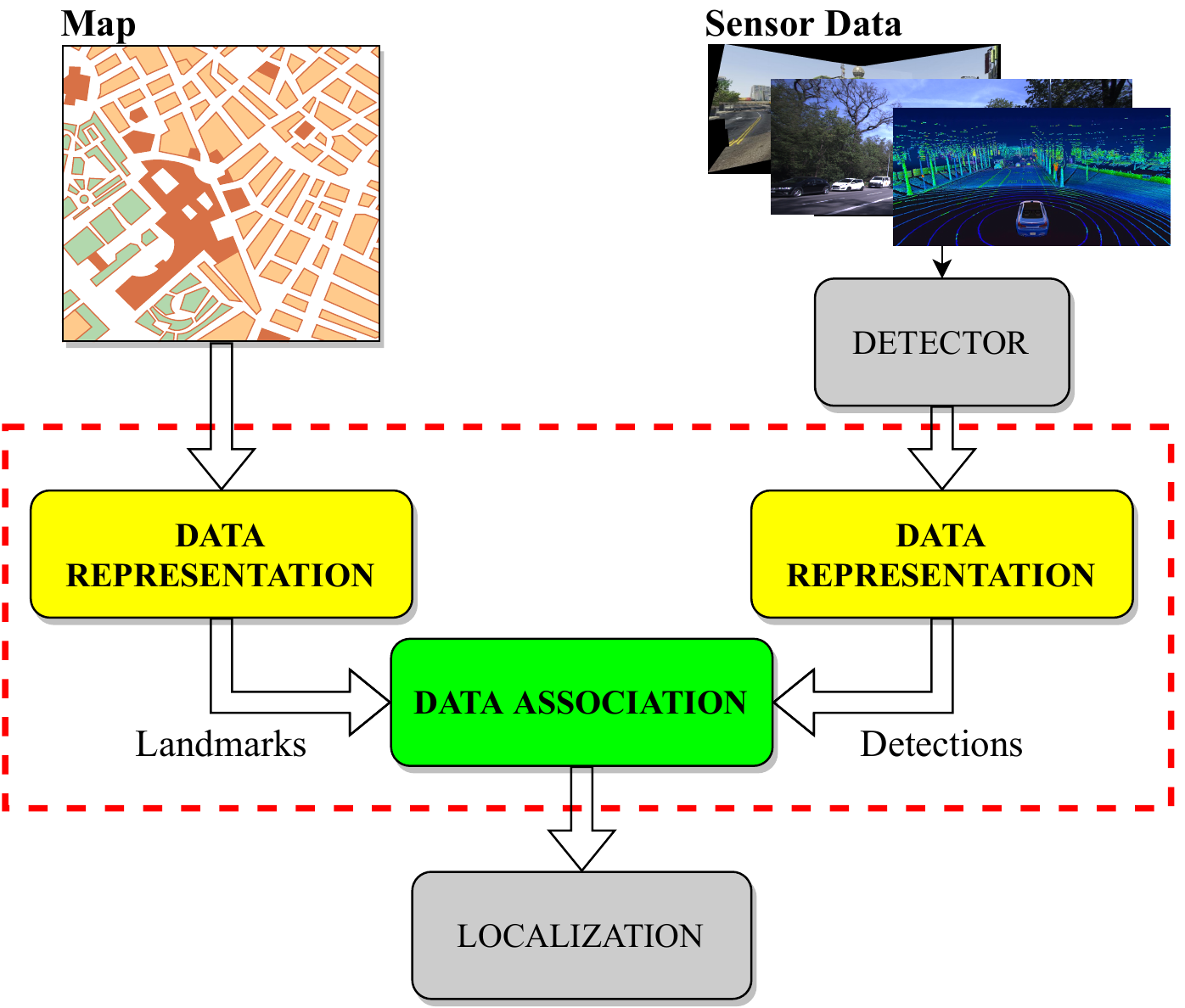}
\caption{An overview of a complete localization system, where the data representation and data association play a key role. Our research focuses on evaluating these modules (inside red dashes) by comparing the presented DA-LMR and the DC-SAC with state-of-the-art approaches.}
\label{fig:overview}
\end{figure}

The landmarks and detections used in the process usually depend on the application and the localization approach. In SLAM or localization in HD maps, lane markings, poles, traffic signs, and traffic lights have been used. While these maps usually provide high local accuracy, they also suffer from drift leading to global inconsistencies due to accumulated errors. To avoid this effect, a so-called geo-referencing is performed, where local detections from sensors are aligned with landmarks from aerial imagery. In that context, traffic signs and lights are unobservable, and poles are hard to detect. Here, lane markings are the potentially best observable landmarks for geo-referencing approaches. These lane markings are usually represented in maps as polylines and dashes.

The question that arises at this point is how to represent polylines and dashes in the data association process. In most works in the literature, there is no evaluation on how lane representations affect the data association, and therefore, the final result of localization. For this reason, in this paper, we focus on evaluating different data association methods using different data representations (yellow and green modules in Fig. \ref{fig:overview}). For this, we propose the \textit{Delta-Angle Lane Marking Representation} (DA-LMR), which represents polylines and dashes as points where an additional dimension represents a proportional value of the differential angle concerning the adjacent line segments. In addition, for data association, we propose a heuristic version of RANSAC, called \textit{Distance-Compatible Sample Consensus} (DC-SAC), that limits possible samples to distance compatible ones.

To summarize, the main contributions are the following:
\begin{itemize}
    \item DA-LMR, a new way of representing lane markings for data association.
    \item A formalization of the data association method DC-SAC, used in our previous work \cite{munoz2020targetless}.
    \item An evaluation of the possible combinations of data representation and data association approaches, including both state-of-the-art, and those proposed in this paper.
\end{itemize}

One of the possible applications of the data representation and data association proposed in this work is the geo-referencing localization using aerial imagery. For this reason, we present this work assuming a 2D world representation.

\section{RELATED WORK}
\label{sec:related_work}
In this section, we briefly review existing data association methods and existing data representations for lane markings.

\subsection{Data Association}
We review the data association methods distinguishing two categories: \textit{pose estimation} and \textit{graph-theoretic}. 

\textit{Pose estimation} methods search for the transformation that minimizes the error between both a set of detections and a set of landmarks or maximizes its number of associations inliers. One of the most widely used \textit{pose estimation} method for data association in localization is Iterative Closest Point (ICP). SLAM approaches such as \cite{weichen2020hector,hu2019mapping,soatti2018implicit}, rely on the ICP for LiDAR scan matching. Currently, state-of-the-art Lidar Odometry And Mapping (LOAM) approaches also implement ICP for the data association process \cite{shan2018lego,shan2020lio}. Nevertheless, this is a local method; hence, it requires a accurate initialization to converge in a global minimum. Further, it is more sensitive to suffer from drift in high outlier scenarios. We do not consider ICP a potential solution for data association as in the context of geo-referencing using aerial imagery no sufficiently accurate pose is available. The random Sample Consensus (RANSAC) method \cite{fischler1981random} is a global data association approach common in the literature. RANSAC follows a strategy that samples hypothesis transformations depending on the data structure. Different authors use RANSAC for localization \cite{yang2010ransac,jiao2020globally} (include aerial imagery-based \cite{hu2019accurate,ramalingam2011pose}), and mapping \cite{cunningham2012fully}.

\textit{Graph-theoretic} methods are based on searching a maximum number of associations being compatible. One common measure of compatibility is Distance Compatibility (DC). This problem is also known in the literature as the maximum clique (MC) problem \cite{wu2015review}. In the past decades, this approach has been used in localization \cite{bailey2000data,mangelson2018pairwise}, including loop closures \cite{frey2019efficient,tian2021kimera}. This is also a global data association approach, and we consider it potentially suitable for localization in aerial imagery. In past years, there has been research focused on a weighed version of the MC problem \cite{wu2012multi,benlic2013breakout}. An interesting variant is CLIPPER \cite{lusk2020clipper}, where the authors implement a weighed MC problem using a continuous relaxation for the optimization process. We can also consider the proposed DC-SAC as a variant of this kind of association.

It is worth noting that there are other data association methods that we don't include in this work because we consider them strongly coupled with the SLAM problem. That is the case of the Hungarian \cite{welte2020improved}, Nearest Neighbor (NN) \cite{castellanos1999spmap}, Joint Compatibility Branch and Bound (JCBB) \cite{neira2001data,yang2020improved}, probabilistic data association based on Mixture Models \cite{zhang2019hierarchical,doherty2020probabilistic}, and Multi-Hypothesis based data association \cite{hsiao2019mh,jiang2021imhs}.

\subsection{Data Representations for polylines}
\label{sec:related_work_dr}
The most common approach presented in the literature to represent data in the data association process is 2D points for 3-DOF localization \cite{heidenreich2015laneslam,vivacqua2017self}. To represent lane markings as 2D points, a previous sampling of polylines becomes necessary. When this process is computed for both landmarks and detections, it is called Point-to-Point (P2P) association. Another strategy is the Point-to-Line association (P2L) \cite{poggenhans2018precise,welte2019estimating}. In this case, the authors project point detections onto the map polylines in each \textit{pose estimation} iteration. However, after projection, the distance is measured between points. Then, we can consider the polylines (landmarks) sampled and converted into 2D points and, roughly speaking, the representation becomes the P2P strategy. Hence, we consider both strategies, P2P and P2L, as unique  2D points-based representations. In other works \cite{hu2019accurate,kummerle2019accurate}, the authors represent the data as lines, also with a previous sampling. In these cases, the authors use a distance metric for line segments, such as Hausdorff distance \cite{yang2021matching,quehl2017good}.

\begin{figure}[t]
\centering
\includegraphics[width=130pt]{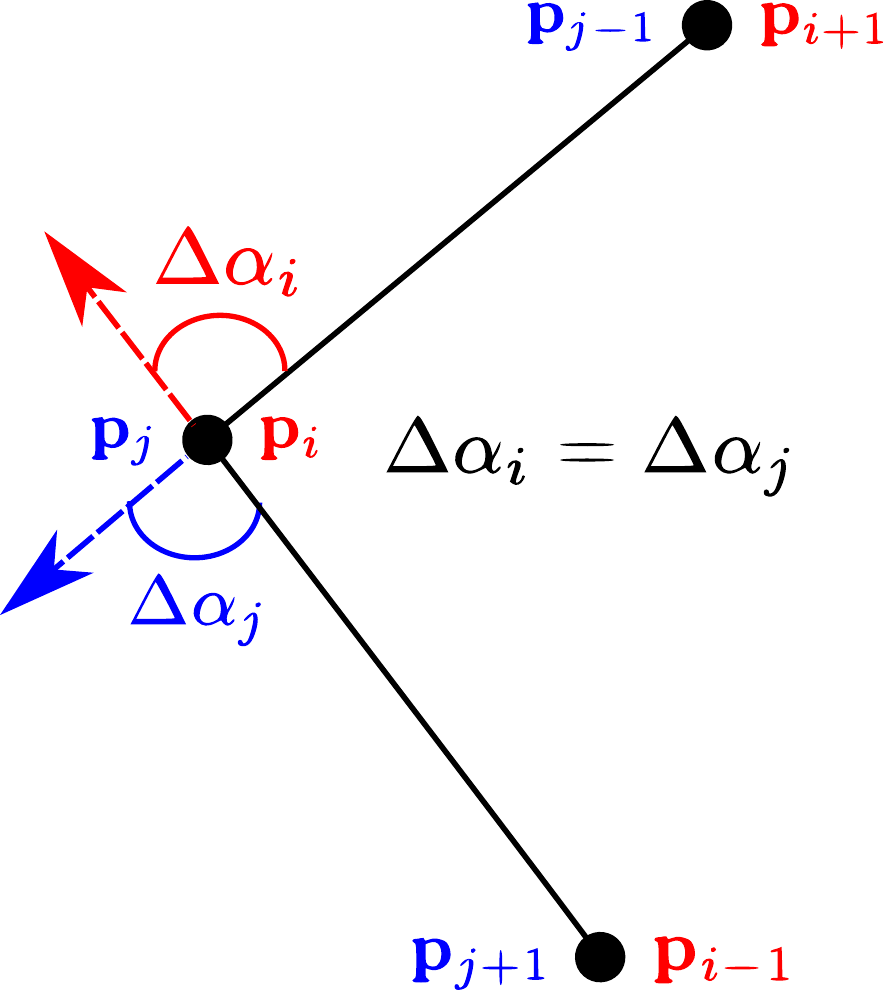}
\caption{Depiction of direction invariance of the delta-angle calculation: in \textit{red}, the $i$-th point, and in \textit{blue}, the $j$-th point are in inverse direction. In both cases, the delta-angle represents the same value.}
\label{fig:delta_angle}
\end{figure}

\section{DELTA ANGLE LANE MARKINGS REPRESENTATION (DA-LMR)}
As discussed in the previous section, the most common representations of lane markings in the process of data association are point-based. Nevertheless, transforming lines and polylines into isolated points entails a loss of information. In a polyline, how the angle of each point changes in a curve provides additional information about the geometric structure of that curve. Following this assumption, we present the \textit{Delta-Angle Lane Marking Representation} (DA-LMR). In the following subsections, we explain how to derive this representation for both polylines and dashes.

\subsection{DA-LMR for polylines}
We can describe a polyline as a set of 2D points $\mathbf{P}^{2D} = (\mathbf{p}^{2D}_0, \mathbf{p}^{2D}_1, ..., \mathbf{p}^{2D}_N)$, where $\mathbf{p}^{2D}_i = (x_i, y_i)$, and where each point has a connection with its adjacent ones. Alternatively, given this connection, we can describe the polyline as a set of vectors $\mathbf{V} = (\vec{\mathbf{v}}_{0}, \vec{\mathbf{v}}_{1}, ..., \vec{\mathbf{v}}_{N-1})$, where each vector is $\vec{\mathbf{v}}_{i} = (x_{i+1} - x_{i}, y_{i+1} - y_{i})$. This vectorial representation provides information about the orientation of each $i$-th point in a polyline. However, representing the orientation in a global reference frame can lead to problems due to the Euler angle wrap-around limitations. To obtain a compact and invariant angle representation, we use the differential angle between adjacent vectors $\vec{\mathbf{v}}_{i-1}$ and $\vec{\mathbf{v}}_{i}$. Then, for each $i$-th point in a polyline, we can obtain the delta-angle as follows:

$$
\Delta\alpha_{i} = \arccos{\left( \frac{\vec{\mathbf{v}}_{i-1} \cdot \vec{\mathbf{v}}_{i}}{\left\lVert \vec{\mathbf{v}}_{i-1}\right\rVert \left\lVert\vec{\mathbf{v}}_{i}\right\rVert} \right)}. 
\eqno{(1)}
$$

It is worth noting, that the points $i = 0$ and $i = N$ does not have adjacent information, and therefore we assign them a default value of $\Delta\alpha_{0} = \Delta\alpha_{N} = 0$. The delta-angle calculation defined in (1) is a non-oriented angle calculation. This entails that the representation is also invariant to the direction in the polyline, i.e., we can sort the polyline from $0$ to $N$ or from $N$ to $0$, obtaining the same result. In Fig. \ref{fig:delta_angle}, we show an example of this direction invariance. In practice, this is important as the representation is independent of the direction from which it is observed.

As the final step of DA-LMR, we use delta-angle information as an extra dimension in the 2D point representation to obtain a compact description suitable for data association methods. Now, the polyline is converted in a set of 3D points $\mathbf{P}^{3D} = (\mathbf{p}^{3D}_0, \mathbf{p}^{3D}_1, ..., \mathbf{p}^{3D}_N)$, where

$$
\mathbf{p}^{3D}_i = (x_i, y_i, \Delta\alpha_i w).
\eqno{(2)}
$$

\begin{figure}[t]
\centering
\includegraphics[width=220pt]{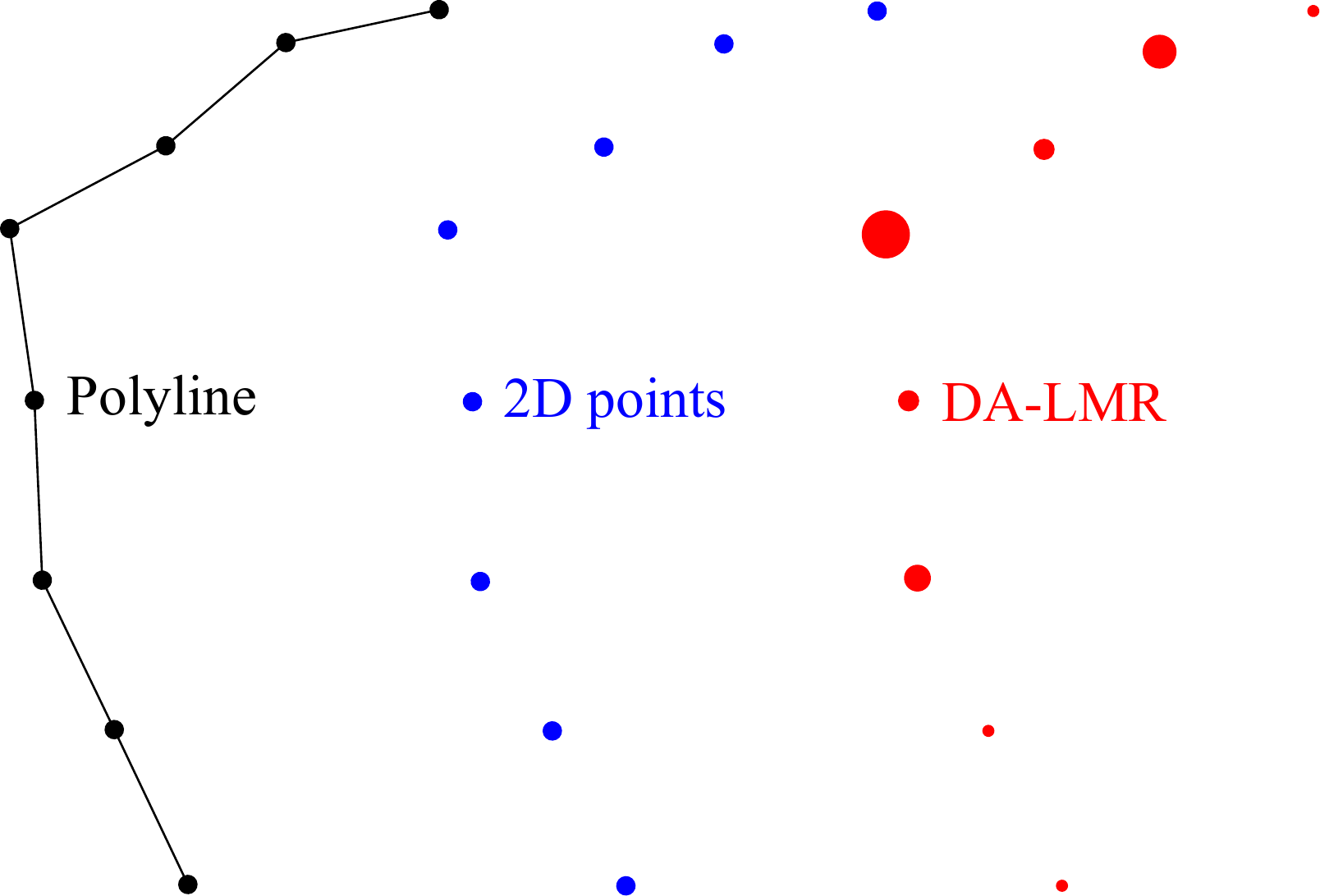}
\caption{Comparison of different representations of the same line: in \textit{black}, line segments representation as polyline, in \textit{blue}, 2D point representation, and in \textit{red}, the proposed DA-LMR. We depict the delta angle for DA-LMR by the size of the circles.}
\label{fig:dalmr_example}
\end{figure}

Here $w$ is a weight that tunes how strongly the delta-angle describes the data structuring. Note that when $w \xrightarrow{} 0$, the representation tends to a traditional 2D representation. In Fig. \ref{fig:dalmr_example}, we show an example of the DA-LMR in contrast to line segments and 2D points. In the case of DA-LMR, we represent the delta angle as the size of the circles. Given this additional dimension, DA-LMR is more informative than previously mentioned representations.

\subsection{DA-LMR for dashes}
For dashed lines, it is not trivial to determine adjacent lines. Hence, for that case, we move the representation formulation to the distance measurement step in the data association process. This distance could be used for the error measurement in \textit{pose estimation} data association, or otherwise, for distance compatibility measurement in \textit{graph-theoretic} data association. 

We can describe dashes as vectors where $\vec{\mathbf{v}}_i = (x_{i_e}-x_{i_s}, y_{i_e}-y_{i_s})$, and where subscripts $i_s$ and $i_e$ denotes start point and end point, respectively for $i$-th vector. Then, given a pair of dashes $\vec{\mathbf{v}}_i$ and $\vec{\mathbf{v}}_j$, we can formulate the distance measurement for DA-LMR as follows:

$$
d_{ij} = \sqrt{(x_j-x_i)^2 + (y_j-y_i)^2 + (\Delta\alpha_{ij} w)^2}.
\eqno{(3)}
$$

Where $(x_i, y_i)$ and $(x_j, y_j)$ are the centroid of each dash respectively. And where

$$
\Delta\alpha_{ij} = \arccos{\left( \frac{\vec{\mathbf{v}}_{i} \cdot \vec{\mathbf{v}}_{j}}{\left\lVert \vec{\mathbf{v}}_{i}\right\rVert \left\lVert\vec{\mathbf{v}}_{j}\right\rVert} \right)}. 
\eqno{(4)}
$$

\begin{figure}[t]
\centering
\includegraphics[width=130pt]{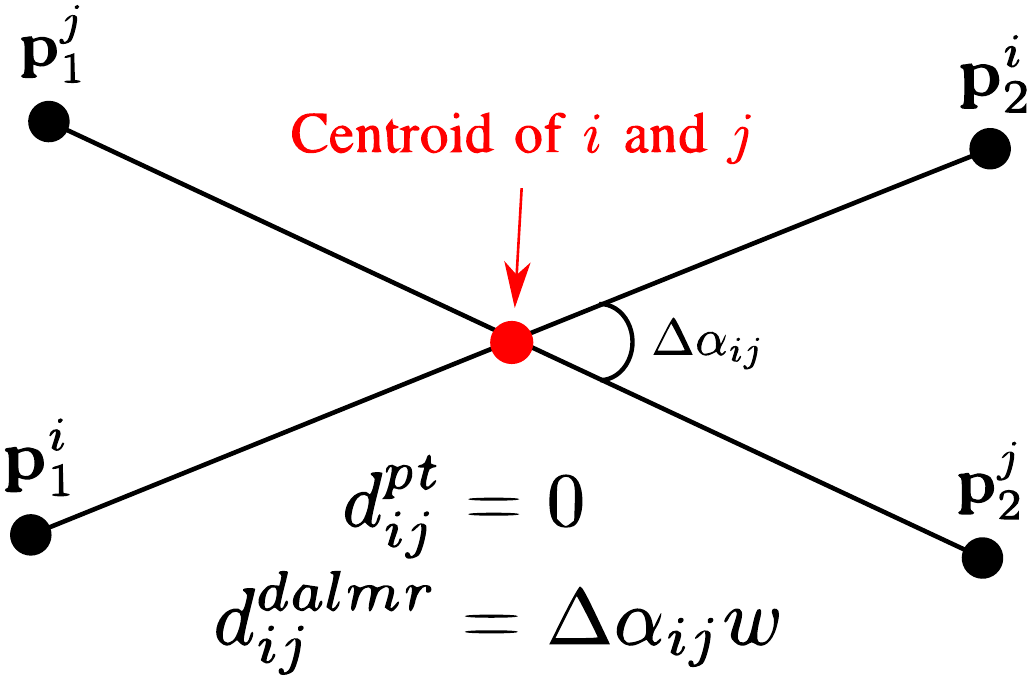}
\caption{Depiction of two dashes intersecting in its centroid. If we compute the distance measurement (3) for a 2D point representation, the result is $d^{pt}_{ij} = 0$. In contrast, if we compute (3) for DA-LMR, the result is $d^{dalmr}_{ij} = \Delta\alpha_{ij} w$, showing the misalignment.}
\label{fig:dalmr_dashes}
\end{figure}

In Fig. \ref{fig:dalmr_dashes}, we show an example of two dashes intersecting in its centroid but misaligned. In this case, if we use a 2D point representation, we obtain a distance measurement $d^{pt}_{ij} = 0$, which doesn't describe the misalignment. Using DA-LMR, we obtain a distance measurement $d^{dalmr}_{ij} = \Delta\alpha_{ij} w$ that uncovers the inconsistency.

\section{DC-SAC DATA ASSOCIATION}
\textit{Distance-Compatible Sample Consensus} (DC-SAC) could be seen as a heuristic version of the RANSAC method. This is a formalization of a previous work \cite{munoz2020targetless} for camera-LiDAR calibration. In \cite{quan2020compatibility}, the authors propose a similar approach but focus on dense point clouds and geometric descriptors.

DA-LMR is a \textit{pose estimation} method that samples the possible transformations. How to sample the data for a $\Delta_T \in SE(2)$ transformation candidate is the key difference between RANSAC and DC-SAC. 

Given a set of landmarks $\mathcal{L}$ and a set of detections $\mathcal{D}$, RANSAC chooses a pair of points randomly from each set. Thereafter, it estimates $\Delta_T \in SE(2)$ by minimizing 

$$
e(\Delta_T) = \left\lVert\mathbf{p}^{d}_1 - \mathbf{p}^{l}_1\right\rVert + \left\lVert\mathbf{p}^{d}_2 - \mathbf{p}^{l}_2\right\rVert.
\eqno{(5)}
$$

Superscripts $l$ and $d$ mean landmarks and detections, respectively. Then, all the points in $\mathcal{D}$ are transformed using the estimated $\Delta_T^*$, obtaining a new set $\mathcal{D}^*$. Thereafter, an error $\epsilon$ is computed by the sum of the distance measurement between each detection in $\mathcal{D}^*$ and the nearest neighbor in $\mathcal{L}$. RANSAC iterates this process obtaining a new $\epsilon$ in each iteration until the error satisfies the condition $\epsilon < \xi$, where $\xi$ is a configurable threshold. The threshold limits the accuracy of the method. For example, a high value for $\xi$ can miss potentially optimal hypotheses producing inaccurate results. In contrast, a low-value configuration of $\xi$ can dramatically increase computation time, possibly making the method unusable.

In the case of DC-SAC, we sample a pair of points in $\mathcal{D}$. However, in contrast to RANSAC, we randomly choose a pair of points in $\mathcal{L}$ that are distance compatible, i.e., that satisfies

$$
\gamma > \left\lvert \left\lVert\mathbf{p}^{d}_1 - \mathbf{p}^{d}_2\right\rVert - \left\lVert\mathbf{p}^{l}_1 - \mathbf{p}^{l}_2\right\rVert \right\rvert.
\eqno{(6)}
$$

Where $\gamma$ is a configurable value. The constraint proposed in (6) could be applied in a general way for data association. Additionally,  as we present this work in localization, we can propose other constraints in the hypothesis space $\mathcal{H}$ depending on the uncertainty of the prior. Given the distribution of $SE(2)$ localization, we can get the standard deviation values for each dimension $(\sigma_x, \sigma_y, \sigma_\theta)$. Then, by considering $\sigma_x$ and $\sigma_y$, we can limit the area of sampling points in $\mathcal{L}$. By using $\sigma_\theta$ and given $\vec{\mathbf{v}}_l = (\mathbf{p}^{l}_1 - \mathbf{p}^{l}_2)$ and $\vec{\mathbf{v}}_d = (\mathbf{p}^{d}_1 - \mathbf{p}^{d}_2)$, we can define an orientation constraint between pairs of samples as follows:

$$
\theta_t > \arccos{\left( \frac{\vec{\mathbf{v}}_{l} \cdot \vec{\mathbf{v}}_{d}}{\left\lVert \vec{\mathbf{v}}_{l}\right\rVert \left\lVert\vec{\mathbf{v}}_{d}\right\rVert} \right)}. 
\eqno{(7)}
$$

Where $\theta_t$ is a configurable value that depend on $\sigma_\theta$. The above-presented heuristic constraints reduce the size of $\mathcal{H}$ dramatically. In this way, we can search the full space $\mathcal{H}$ for the minimum value of $\epsilon$ instead of iterating until $\epsilon < \xi$. This is the best scenario for converging to a potentially optimal solution. In the next section, we demonstrate experimentally how DC-SAC produces more accurate results than RANSAC by configuring a value of $\xi$ where both methods run with similar computation time.

\begin{figure}[t]
    \centering
    \begin{tikzpicture}
    \begin{axis}[
    name=plot,
    xlabel={$\sigma$ (m)}, ylabel={\%},
    xmin=0.1, xmax = 0.5, ymin=70, ymax=100,
    width=230pt,height=200pt,
    grid=both,
    grid style={line width=.1pt, draw=gray!10},
    major grid style={line width=.2pt,draw=gray!50},
    legend pos=south west,
    legend style={nodes={scale=0.5, transform shape}}
    ]
    
    \addplot[color={rgb:red,4;green,0;yellow,5},mark=pentagon*,very thick] table{./data/05_clipper_lsr_f1.txt};
    \addlegendentry{LSR (F1 score)}
    \addplot[color={rgb:red,4;green,0;yellow,5},dotted,very thick] table{./data/05_clipper_lsr_p.txt};
    \addlegendentry{LSR (Precision)}
    \addplot[color={rgb:red,4;green,0;yellow,5},dashed,very thick] table{./data/05_clipper_lsr_r.txt};
    \addlegendentry{LSR (Recall)}

    \addplot[color={rgb:red,1;green,2;blue,5},mark=triangle*,very thick] table{./data/05_clipper_2dpr_f1.txt};
    \addlegendentry{PR (F1 score)}
    \addplot[color={rgb:red,1;green,2;blue,5},dotted,very thick] table{./data/05_clipper_2dpr_p.txt};
    \addlegendentry{PR (Precision)}
    \addplot[color={rgb:red,1;green,2;blue,5},dashed,very thick] table{./data/05_clipper_2dpr_r.txt};
    \addlegendentry{PR (Recall)}

    \addplot[color={rgb:red,4;green,0;yellow,1},mark=square*,very thick] table{./data/05_clipper_dalmr_f1.txt};
    \addlegendentry{DA-LMR (F1 score)}
    \addplot[color={rgb:red,4;green,0;yellow,1},dotted,very thick] table{./data/05_clipper_dalmr_p.txt};
    \addlegendentry{DA-LMR (Precision)}
    \addplot[color={rgb:red,4;green,0;yellow,1},dashed,very thick] table{./data/05_clipper_dalmr_r.txt};
    \addlegendentry{DA-LMR (Recall)}

    \end{axis}
    \end{tikzpicture}
    \caption{Comparison in F1 score (\textit{solid}), precision (\textit{dotted}) and recall (\textit{dashed}) metrics between LSR, PR, and DA-LMR for CLIPPER \cite{lusk2020clipper} as data association method.}
    \label{fig:clipper_p}
\end{figure}
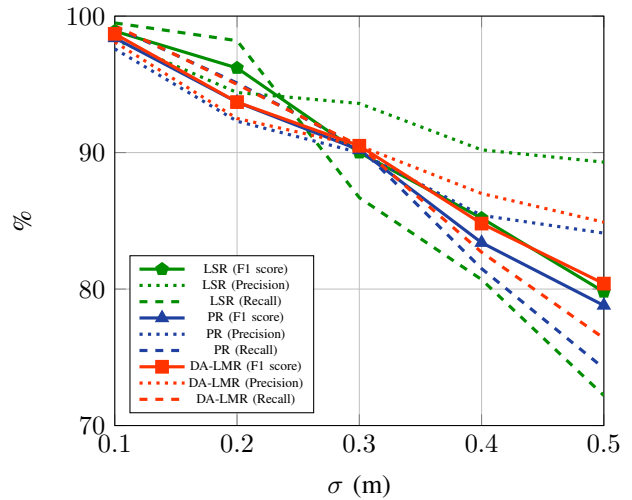

\begin{figure}[t]
    \centering
    \begin{tikzpicture}
    \begin{axis}[
    name=plot,
    xlabel={$\sigma$ (m)}, ylabel={\%},
    xmin=0.1, xmax = 0.5, ymin=70, ymax=100,
    width=230pt,height=200pt,
    grid=both,
    grid style={line width=.1pt, draw=gray!10},
    major grid style={line width=.2pt,draw=gray!50},
    legend pos=south west,
    legend style={nodes={scale=0.5, transform shape}}
    ]
    
    \addplot[color={rgb:red,4;green,0;yellow,5},mark=pentagon*,very thick] table{./data/06_mc_lsr_f1.txt};
    \addlegendentry{LSR (F1 score)}
    \addplot[color={rgb:red,4;green,0;yellow,5},dotted,very thick] table{./data/06_mc_lsr_p.txt};
    \addlegendentry{LSR (Precision)}
    \addplot[color={rgb:red,4;green,0;yellow,5},dashed,very thick] table{./data/06_mc_lsr_r.txt};
    \addlegendentry{LSR (Recall)}

    \addplot[color={rgb:red,1;green,2;blue,5},mark=triangle*,very thick] table{./data/06_mc_2dpr_f1.txt};
    \addlegendentry{PR (F1 score)}
    \addplot[color={rgb:red,1;green,2;blue,5},dotted,very thick] table{./data/06_mc_2dpr_p.txt};
    \addlegendentry{PR (Precision)}
    \addplot[color={rgb:red,1;green,2;blue,5},dashed,very thick] table{./data/06_mc_2dpr_r.txt};
    \addlegendentry{PR (Recall)}

    \addplot[color={rgb:red,4;green,0;yellow,1},mark=square*,very thick] table{./data/06_mc_dalmr_f1.txt};
    \addlegendentry{DA-LMR (F1 score)}
    \addplot[color={rgb:red,4;green,0;yellow,1},dotted,very thick] table{./data/06_mc_dalmr_p.txt};
    \addlegendentry{DA-LMR (Precision)}
    \addplot[color={rgb:red,4;green,0;yellow,1},dashed,very thick] table{./data/06_mc_dalmr_r.txt};
    \addlegendentry{DA-LMR (Recall)}

    \end{axis}
    \end{tikzpicture}
    \caption{Comparison in F1 score (\textit{solid}), precision (\textit{dotted}) and recall (\textit{dashed}) metrics between LSR, PR, and DA-LMR for Max Clique \cite{konc2007improved} as data association method.}
    \label{fig:mc_p}
\end{figure}
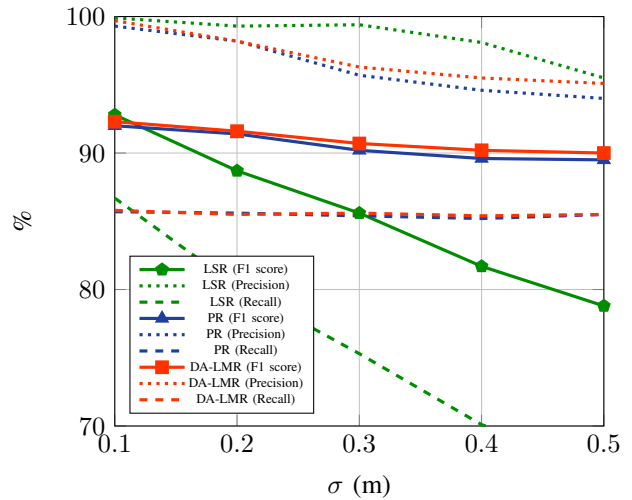

\section{EVALUATION}

The evaluation step aims to compare the presented DA-LMR with other data representations as well as the presented DC-SAC with other data association methods. It is difficult to perform a quantitative comparison using maps and real onboard sensor data since it is hard to obtain ground truth. Hence, in Section \ref{sec:quantitative}, we generate synthetic detections $\mathcal{D}^s$ from the original landmarks $\mathcal{L}$ of a real map by cropping, transforming, and adding noise. In Section \ref{sec:qualitative}, we show qualitative results using the previously used $\mathcal{L}$ and real detections $\mathcal{D}^r$ from stereo cameras in a real autonomous vehicle.

\subsection{Quantitative results}
\label{sec:quantitative}

For quantitative evaluation, we use a map that contains continuous and dashed lane markings. This map was labeled by hand from aerial imagery from the city of Karlsruhe, Germany. We define the set of landmarks $\mathcal{L}$ by sampling the map polylines with steps of $\SI{1}{\meter}$, and we use this labeled data as ground truth. Then, using a pre-recorded trajectory in the mapped area as a reference, we apply a sliding window to $\mathcal{L}$. In each iteration, we crop a set of data that forms a new  $\mathcal{D}^s_i$, where the subscript $i$ indicates the iteration of the sliding window, and the superscript $s$ denotes synthetic. Thereafter, we apply a random 3-DOF rigid body transformation $\mathbf{t} = (t_x, t_y)$ and $\mathbf{R} = f(r_\theta)$ to each $\mathcal{D}^s_i$. Where $t_x$ and $t_y$ are uniformly distributed random variables with range $(\SI{-5}{\meter}, \SI{5}{\meter})$, and where $r_\theta$ is also uniformly distributed random variable with range $(\SI{-5}{\degree}, \SI{5}{\degree})$. Also, for each experiment, we produce additive Gaussian noise with a range of $\sigma$ between $\SI{0.1}{\meter}$ and $\SI{0.5}{\meter}$ with steps of $\SI{0.1}{\meter}$. Finally, we include outliers, where the number of outliers is $\SI{10}{\percent}$ of the samples in $\mathcal{D}^s_i$. It is worth noting that it is not recommendable to execute the data association process in the complete set $\mathcal{L}$. Hence, assuming we have a prior localization, we generate a set $\mathcal{L}_i$ bigger than $\mathcal{D}^s_i$ in each iteration of the sliding window that covers an area around it.

\begin{figure}[t]
    \centering
    \begin{tikzpicture}
    \begin{axis}[
    name=plot,
    xlabel={$\sigma$ (m)}, ylabel={\%},
    xmin=0.1, xmax = 0.5, ymin=90, ymax=100,
    width=230pt,height=195pt,
    grid=both,
    grid style={line width=.1pt, draw=gray!10},
    major grid style={line width=.2pt,draw=gray!50},
    legend pos=south west,
    legend style={nodes={scale=0.5, transform shape}}
    ]
    
    \addplot[color={rgb:red,4;green,0;yellow,5},mark=pentagon*,very thick] table{./data/07_ransac_lsr_f1.txt};
    \addlegendentry{LSR (F1 score)}
    \addplot[color={rgb:red,4;green,0;yellow,5},dotted,very thick] table{./data/07_ransac_lsr_p.txt};
    \addlegendentry{LSR (Precision)}
    \addplot[color={rgb:red,4;green,0;yellow,5},dashed,very thick] table{./data/07_ransac_lsr_r.txt};
    \addlegendentry{LSR (Recall)}

    \addplot[color={rgb:red,1;green,2;blue,5},mark=triangle*,very thick] table{./data/07_ransac_2dpr_f1.txt};
    \addlegendentry{PR (F1 score)}
    \addplot[color={rgb:red,1;green,2;blue,5},dotted,very thick] table{./data/07_ransac_2dpr_p.txt};
    \addlegendentry{PR (Precision)}
    \addplot[color={rgb:red,1;green,2;blue,5},dashed,very thick] table{./data/07_ransac_2dpr_r.txt};
    \addlegendentry{PR (Recall)}

    \addplot[color={rgb:red,4;green,0;yellow,1},mark=square*,very thick] table{./data/07_ransac_dalmr_f1.txt};
    \addlegendentry{DA-LMR (F1 score)}
    \addplot[color={rgb:red,4;green,0;yellow,1},dotted,very thick] table{./data/07_ransac_dalmr_p.txt};
    \addlegendentry{DA-LMR (Precision)}
    \addplot[color={rgb:red,4;green,0;yellow,1},dashed,very thick] table{./data/07_ransac_dalmr_r.txt};
    \addlegendentry{DA-LMR (Recall)}

    \end{axis}
    \end{tikzpicture}
    \caption{Comparison in F1 score (\textit{solid}), precision (\textit{dotted}) and recall (\textit{dashed}) metrics between LSR, PR, and DA-LMR for RANSAC \cite{hu2019accurate} as data association method.}
    \label{fig:ransac_p}
\end{figure}
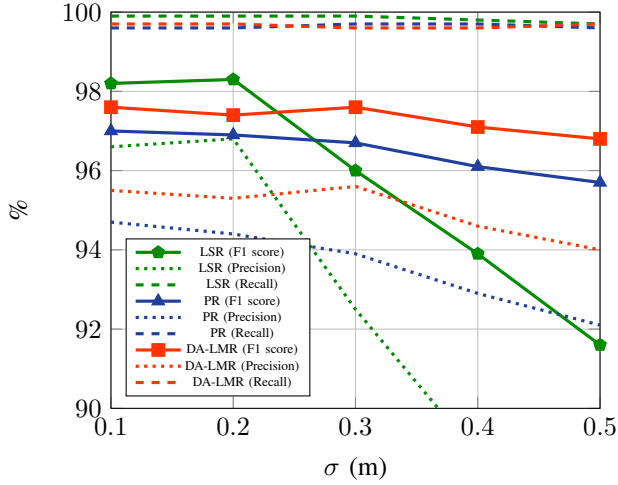

\begin{figure}[t]
    \centering
    \begin{tikzpicture}
    \begin{axis}[
    name=plot,
    xlabel={$\sigma$ (m)}, ylabel={\%},
    xmin=0.1, xmax = 0.5, ymin=90, ymax=100,
    width=230pt,height=195pt,
    grid=both,
    grid style={line width=.1pt, draw=gray!10},
    major grid style={line width=.2pt,draw=gray!50},
    legend pos=south west,
    legend style={nodes={scale=0.5, transform shape}}
    ]
    
    \addplot[color={rgb:red,4;green,0;yellow,5},mark=pentagon*,very thick] table{./data/08_dcsac_lsr_f1.txt};
    \addlegendentry{LSR (F1 score)}
    \addplot[color={rgb:red,4;green,0;yellow,5},dotted,very thick] table{./data/08_dcsac_lsr_p.txt};
    \addlegendentry{LSR (Precision)}
    \addplot[color={rgb:red,4;green,0;yellow,5},dashed,very thick] table{./data/08_dcsac_lsr_r.txt};
    \addlegendentry{LSR (Recall)}

    \addplot[color={rgb:red,1;green,2;blue,5},mark=triangle*,very thick] table{./data/08_dcsac_2dpr_f1.txt};
    \addlegendentry{PR (F1 score)}
    \addplot[color={rgb:red,1;green,2;blue,5},dotted,very thick] table{./data/08_dcsac_2dpr_p.txt};
    \addlegendentry{PR (Precision)}
    \addplot[color={rgb:red,1;green,2;blue,5},dashed,very thick] table{./data/08_dcsac_2dpr_r.txt};
    \addlegendentry{PR (Recall)}

    \addplot[color={rgb:red,4;green,0;yellow,1},mark=square*,very thick] table{./data/08_dcsac_dalmr_f1.txt};
    \addlegendentry{DA-LMR (F1 score)}
    \addplot[color={rgb:red,4;green,0;yellow,1},dotted,very thick] table{./data/08_dcsac_dalmr_p.txt};
    \addlegendentry{DA-LMR (Precision)}
    \addplot[color={rgb:red,4;green,0;yellow,1},dashed,very thick] table{./data/08_dcsac_dalmr_r.txt};
    \addlegendentry{DA-LMR (Recall)}

    \end{axis}
    \end{tikzpicture}
    \caption{Comparison in F1 score (\textit{solid}), precision (\textit{dotted}) and recall (\textit{dashed}) metrics between LSR, PR, and DA-LMR for the proposed DC-SAC as data association method.}
    \label{fig:dcsac_p}
\end{figure}
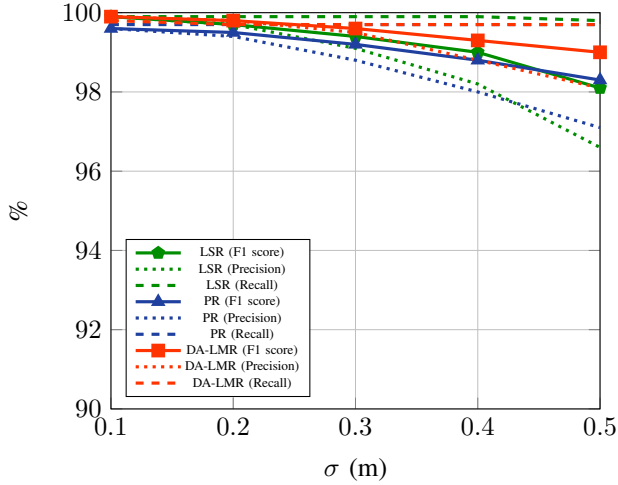

\begin{figure*}[t]
\centering
\includegraphics[width=500pt]{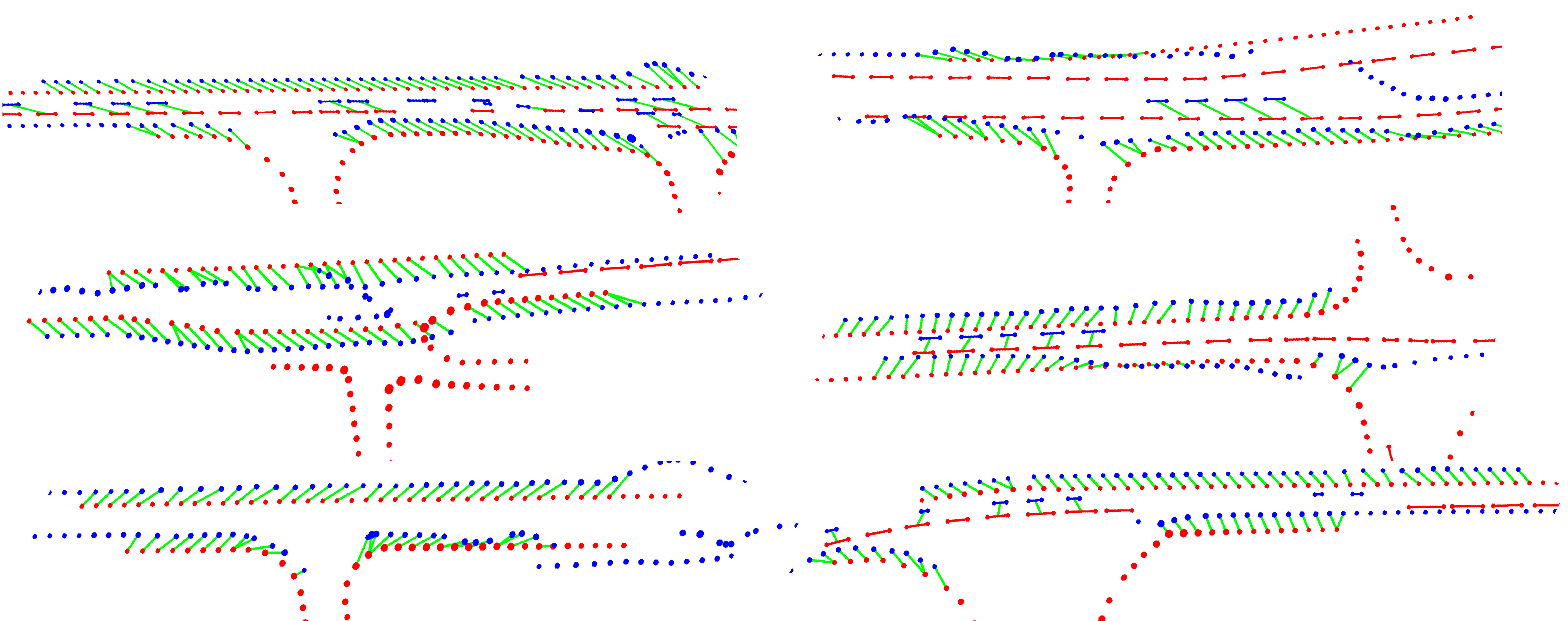}
\caption{Examples of data association results. In \textit{red}, we show the $\mathcal{L}_i$, while in \textit{blue}, we show $\mathcal{D}^r_i$. \textit{Green} lines indicate the association between landmarks and detections. The size of circles is proportional to the Z-axis value.}
\label{fig:mosaic}
\end{figure*}

At this point, given $\mathcal{L}_i$ and $\mathcal{D}^s_i$, we run a data association process for each data representation and each $\sigma$ value of the range mentioned above. We use the representations shown Fig. \ref{fig:dalmr_example}, i.e., line segments representation (LSR), 2D points representation (PR), and the proposed DA-LMR with $w = \SI{5}{\frac{\meter}{\radian}}$. For metrics, we use Euclidean distance for PR and DA-LMR and Hausdorff distance for LSR. Finally, we repeat the process for the different data association methods: CLIPPER \cite{lusk2020clipper}, Max Clique \cite{konc2007improved}, RANSAC \cite{hu2019accurate}, and the proposed DC-SAC.

In Fig. \ref{fig:clipper_p}, we show a comparison in F1 score, precision and recall metrics between the representation approaches using CLIPPER \cite{lusk2020clipper} as data association method. The statistics are accumulated for all $i$ values, i.e., for the complete pre-recorded trajectory. For this experiment, we configure the distance compatibility (6) parameter as $\gamma = 3 \sigma$\footnote{It is worth noting that the parameter $\gamma$ only defined for DC-SAC is also used in graph-theoretic approaches because such methods are based on distance compatibility constraints too.}. As we can see in Fig. \ref{fig:clipper_p}, DA-LMR improves the results of PR for both precision and recall. LSR shows the best results in precision. However, it generates the worst results in recall. In other words, LSR can associate less data than the other, but with better precision. 

In the case of Max Clique \cite{konc2007improved}, for this experiment, we also configure the distance compatibility (6) parameter as $\gamma = 3 \sigma$. We can see in Fig. \ref{fig:mc_p} that the behavior of the different representations is similar to the previous. But in this case, LSR seems more sensitive against the noise in recall statistics than PR and DA-LMR.

For RANSAC \cite{hu2019accurate}, we configure $\xi$ with a value that produces a computational time similar to the DC-SAC method. The results for RANSAC are strongly correlated with the threshold $\xi$. For this reason, the results are stable against the noise. We can see this behavior in Fig. \ref{fig:ransac_p} for PR and DA-LMR, where DA-LMR improves the results of PR. However, LSR shows different behavior and presents less stability against noise again.

In Fig. \ref{fig:dcsac_p}, we compare the data representation approaches using the presented DC-SAC data association method and configure $\gamma = 3 \sigma$. In this case, we can see the best results in precision with DA-LMR. While in recall, LSR shows the best results, but the difference is relatively small.

In general, LSR shows good results in low-noise scenarios, but it is more sensitive to noise disturbances. The PR and DA-LMR show more robustness against noise, and DA-LMR improves the results of PR in all cases.

Comparing the data association methods, we found the \textit{graph-theoretic} ones (CLIPPER and Max Clique) more sensitive against noise. Especially the CLIPPER method. Max Clique shows good results for low-noise, but for high noise, the precision is similar to RANSAC and the recall much worst. In contrast, we found the \textit{pose estimation} methods more robust against noise. RANSAC shows reasonably good results. In precision it is similar to the worst case of Max Clique, but in recall shows more stability. DC-SAC shows, in general, the best results in precision and recall and, like RANSAC, high robustness against noise.

Finally, in Table \ref{tab:time_consuming}, we show the evaluation of the computational time for each data association and each data representation depending on the samples number $\left\lvert \mathcal{L}_i \right\rvert + \left\lvert \mathcal{D}^s_i \right\rvert$. The experiments were performed on an i7-7700HQ CPU with 16 GB of RAM in a C++ compiled code. We can see that RANSAC and DC-SAC show similar results for the configuration described in this section and are reasonable. In the case of Max Clique, it improves the time consumption for low sample scenarios, but for $>200$ samples, it is slower than SAC-based methods. The CLIPPER method is the slowest of all. For 400 samples, the CPU cannot even run a complete process. Comparing data representation, LSR shows slower behavior than PR and DA-LMR for SAC-based methods. However, for \textit{graph-theoretic} approaches, the behavior is different. In that case, LSR is faster than the others representations. In all cases, DA-LMR is faster than PR.

\begin{table}[ht]
\caption{Computation time evaluation in seconds.}
\label{tab:time_consuming}
\begin{center}
\begin{tabular}{c c c c c c c}
\hline
Samples & DR & \textbf{DC-SAC} & \textbf{RANSAC} & \textbf{MC} & \textbf{CLIPPER} \\
\hline
\hline
100 & LSR & 0.09 & 0.06 & 0.02 & 0.17 \\
    & PR & 0.06 & 0.06 & 0.03 & 0.69 \\
    & DA-LMR & 0.06 & 0.06 & 0.03 & 0.31 \\
\hline
200 & LSR & 0.91 & 1.12 & 0.27 & 4.79 \\
    & PR & 0.67 & 0.69 & 0.53 & 21.16 \\
    & DA-LMR & 0.61 & 0.59 & 0.55 & 10.50 \\
\hline
300 & LSR & 4.58 & 5.15 & 2.86 & 41.32 \\
    & PR & 3.27 & 2.97 & 4.86 & 151.3 \\
    & DA-LMR & 2.91 & 2.83 & 4.35 & 55.62 \\
\hline
400 & LSR & 18.55 & 19.72 & 12.77 & - \\
    & PR & 12.88 & 13.05 & 18.61 & - \\
    & DA-LMR & 11.91 & 12.64 & 18.53 & - \\
\hline
\end{tabular}
\end{center}
\end{table}

As we obtain the highest values on precision and recall, we consider that the combination of the proposed data representation approach DA-LMR, and the presented data association DC-SAC, is the best combination among the evaluated for a localization application and specially for geo-referencing using aerial imagery.

\subsection{Qualitative results}
\label{sec:qualitative}
In Section \ref{sec:quantitative}, we compared different data association methods with different data representations, and we concluded that DA-LMR and DC-SAC are the preferable combination. In this section, we show some qualitative results with the chosen combination. For the evaluation, we use the same map as in the previous section and the same pre-recorded trajectory. We use the same $\mathcal{L}_i$ as a set of landmarks for each sliding window iteration. However, in this case, we use real data for detections ($\mathcal{D}^r_i$) obtained while driving the pre-recorded trajectory. Detections were obtained using stereo cameras from our
experimental vehicle \textit{BerthaOne} \cite{tacs2017making}. Then, given $\mathcal{L}_i$ and $\mathcal{D}^r_i$, we iteratively run the process of DA-LMR and DC-SAC for the complete trajectory.

In Fig. \ref{fig:mosaic}, we show some examples of data association results. In \textit{red}, we show the $\mathcal{L}_i$, while in \textit{blue}, we show $\mathcal{D}^r_i$. \textit{Green} lines indicate the association between landmarks and detections. The size of circles is proportional to the delta angles. The results look very reasonable for its implementation in a complete localization approach.

\section{CONCLUSIONS}
We proposed DA-LMR, a robust data representation for lane markings in data association processes for localization. As we demonstrated experimentally, this representation provides richer information of geometrical data structure than others, such as line segments and 2D points. Furthermore, we proposed DC-SAC, a data association method that can improve the results by eliminating potentially non-optimal hypotheses in the space using distance compatibility constraints. We also demonstrate experimentally how the presented method improves the results of other state-of-the-art ones, such as RANSAC, Max Clique, and CLIPPER.

In future work, we will apply the combination of DA-LMR and DC-SAC for a geo-referencing localization using aerial imagery. Although we have emphasized this kind of localization in this work, we also plan to apply the presented method to localization in HD maps.


\bibliography{references.bib}{}
\bibliographystyle{IEEEtran}

\end{document}